\pdfoutput=1

\documentclass[11pt]{article}
\PassOptionsToPackage{table,xcdraw}{xcolor}
\usepackage[preprint]{acl}
\usepackage{times}
\usepackage{latexsym}

\usepackage[T1]{fontenc}

\usepackage[utf8]{inputenc}
\usepackage{microtype}

\usepackage{inconsolata}

\usepackage{graphicx}

\usepackage{xcolor}
\usepackage{geometry}
\usepackage[raster, most]{tcolorbox}
\usepackage{array}

\usepackage{enumitem}

\title{How LLMs Comprehend Temporal Meaning in Narratives: 

A Case Study in Cognitive Evaluation of LLMs}

\author{
 \textbf{Karin de Langis\textsuperscript{1}},
 \textbf{Jong Inn Park\textsuperscript{1}},
 \textbf{Andreas Schramm\textsuperscript{2}},
 \textbf{Bin Hu\textsuperscript{1}},
 \\
 \textbf{Khanh Chi Le\textsuperscript{1}},
 \textbf{Michael Mensink\textsuperscript{3}},
 \textbf{Ahn Thu Tong\textsuperscript{2}},
 \textbf{Dongyeop Kang\textsuperscript{1}}
\\
 \textsuperscript{1}University of Minnesota,
 \textsuperscript{2}Hamline University,
 \textsuperscript{3}University of Wisconsin-Stout
\\
 \small{
   \textbf{Correspondence:} dento019@umn.edu
 }
}

\begin{document}
\maketitle
\begin{abstract}
Large language models (LLMs) exihibit increasingly sophisticated linguistic capabilities, yet the extent to which these behaviors reflect human-like cognition versus advanced pattern recognition remains an open question.
In this study, we investigate how LLMs process the temporal meaning of linguistic aspect in narratives that were previously used in human studies. Using an Expert-in-the-Loop probing pipeline, we conduct a series of targeted experiments to assess whether LLMs construct semantic representations and pragmatic inferences in a human-like manner.
Our findings show that LLMs over-rely on prototypicality, produce inconsistent aspectual judgments, and struggle with causal reasoning derived from aspect, raising concerns about their ability to fully comprehend narratives.
These results suggest that LLMs process aspect fundamentally differently from humans and lack robust narrative understanding.
Beyond these empirical findings, we develop a standardized experimental framework for the reliable  assessment of LLMs' cognitive and linguistic capabilities.
\end{abstract}

\section{Introduction}

\begin{figure}
    \centering
    \includegraphics[width=1\linewidth]{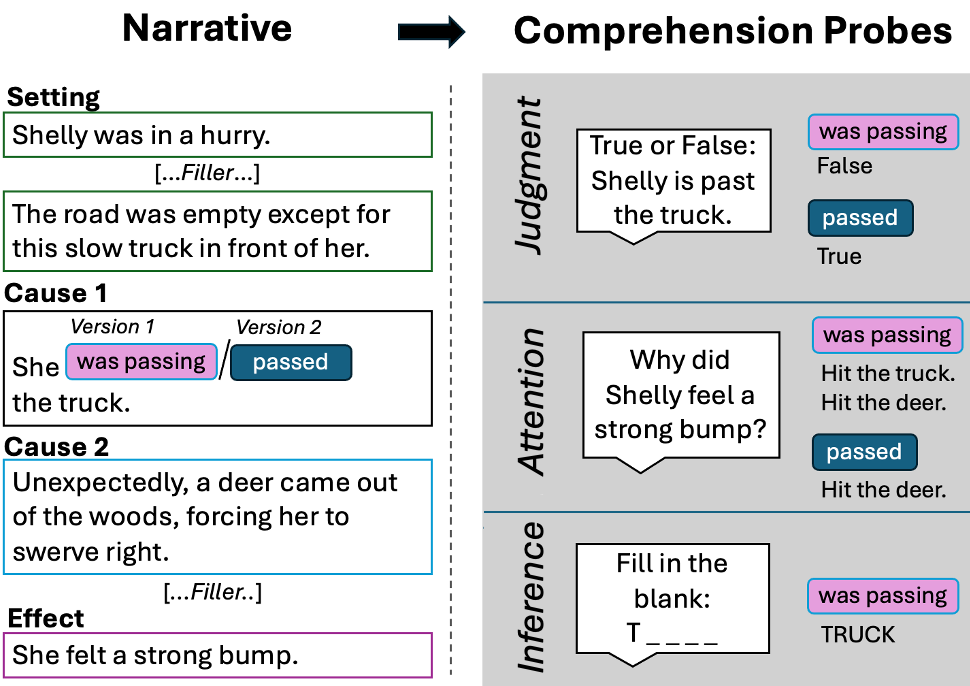}
    \caption{We examine how LLMs understand differences in aspect by presenting LLMs with narratives that have a key word either in the imperfective or perfective (e.g., ``was passing'' vs. ``passed'') followed by comprehension probes adapted from previous human studies.}
    \label{fig:narrative}
\end{figure}

Advanced state-of-the-art large language models (LLMs) exhibit strikingly human-like behavior. However, the extent to which these models demonstrate genuine understanding, rather than sophisticated statistical pattern matching, remains an open question \cite{maleki2024ai, li2023dark, bender2021dangers,das2024surface}. A growing body of research attempts to address this issue by applying experimental methodologies from cognitive science to evaluate cognitive and linguistic capabilities of LLMs \cite{ivanova2025evaluate}. We contribute to this effort by investigating how LLMs comprehend the semantic and pragmatic temporal meanings expressed by \textit{aspect} in narratives (e.g., Fig.~\ref{fig:narrative}).

Aspect is a complex two-component linguistic phenomenon that allows the speaker to present the inherent temporal structure of a given event (known as lexical aspect\footnote{For instance, the lexical aspect of ``walk in the park'' and ``walk to the park'' differ, since ``walk to the park'' has a natural endpoint and ``walk in the park'' does not.}) with different focuses (referred to as grammatical aspect\footnote{The two main grammatical aspects in English are the perfective (e.g., ``walked'') and the imperfective (e.g., ``was walking'').}). 
Crucially, the imperfective aspect (e.g., “was passing”) presents an event as ongoing, without reaching a final state. In narrative contexts, this has been shown to \textit{facilitate causal inference}, as ongoing events are more likely to be retained in working memory and integrated into a reader’s mental model of the story \cite{magliano2000verb, mozuraitis2013younger, schramm2016processing}. 

\begin{figure}[t]
    \centering
    \includegraphics[width=\linewidth]{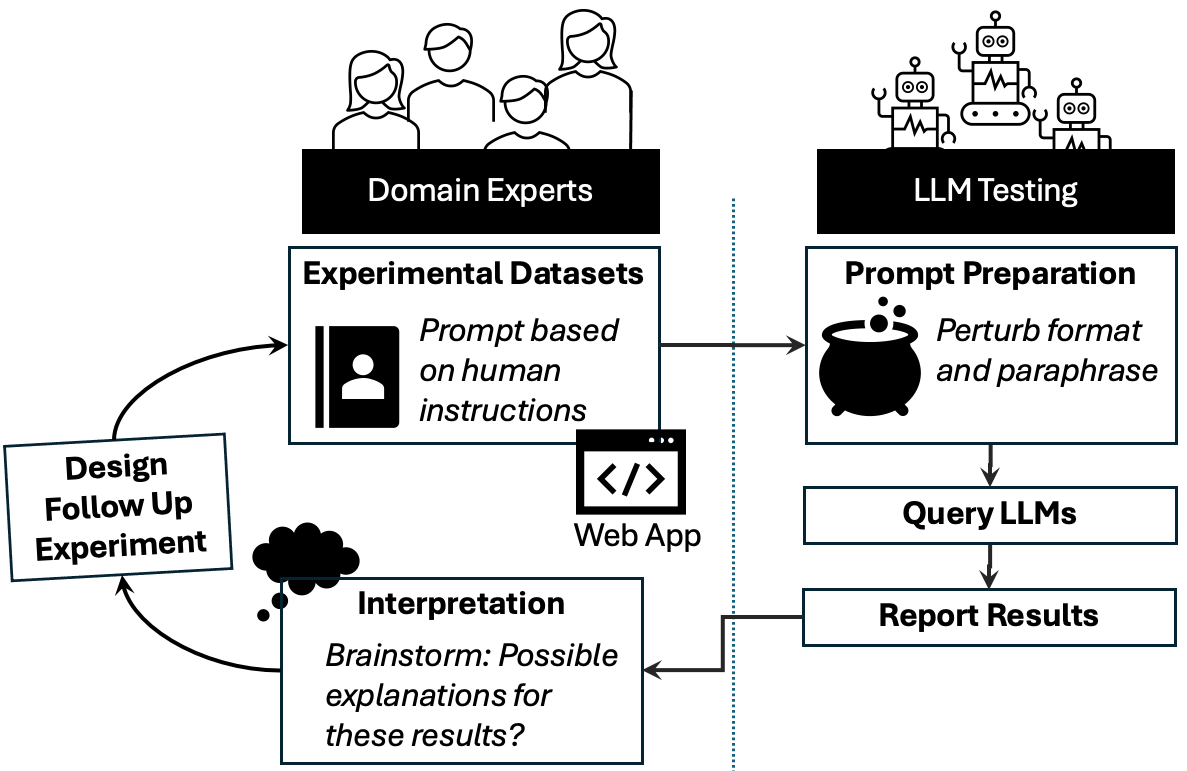}
    \caption{A conceptual overview of our Expert-in-the-Loop probing pipeline for assessing cognitive abilities of LLMs, designed in close collaboration with domain experts from cognitive science.} 
    \label{fig:overview}
\end{figure}

Because of its nuanced role in human language processing, aspect provides a valuable test case for probing LLM capabilities in semantic and pragmatic comprehension.
Do LLMs, like humans, recognize aspect as a cue for retaining an event as open-ended or closing it off as completed? Do they grasp the causal implications conveyed by aspectual distinctions, and can they recognize the non-prototypical use of an open-ended event in a narrative?
To answer these questions, we conduct a series of experiments using narrative stimuli and experimental paradigms previously developed by cognitive linguists to study similar questions in humans. To support these experiments, we build an Expert-in-the-Loop probing pipeline for LLM assessments based on experimental materials from existing studies (Fig. ~\ref{fig:overview}). 

Our results indicate that while LLMs have some understanding of aspectual semantics, their comprehension remains weaker than expected and their cognitive performance across various measures is inconsistent (i.e., no model maintains human-like responses across all measures). Moreover, we find that many state-of-the-art LLMs struggle with the pragmatic causal narrative and backgrounding implications of imperfective aspect, suggesting an over-reliance on prototypical linguistic structures rather than a flexible, context-driven interpretation. These findings contribute to the growing understanding of LLM cognition \citep{apidianaki2024language, ohmer2024form} and pragmatic reasoning \citep{beuls2024humans, sravanthi2024pub}, underscoring the limitations of current LLMs and highlighting the need for further research into the cognitive foundations of LLM behavior. 

\section{Related Work}
Recent advances in LLM have led to an emerging area of research that assesses abilities and characteristics of LLMs by leveraging datasets and methods from human studies \citep{ivanova2025evaluate, hagendorff2023human}. These works study a variety of domains: cognition \citep{roberts2024using, echterhoff-etal-2024-cognitive}, causal reasoning \citep{dasgupta2022language, binz2023using}, decision-making \citep{hagendorff2023human, pmlr-v235-coda-forno24a}, philosophy of mind \citep{ullman2023large, echterhoff-etal-2024-cognitive}, psycholinguistics \citep{bazhukov-etal-2024-models, lee-etal-2024-psycholinguistic}, and emotion and personality \citep{coda2023inducing, huang2024on}. 


\paragraph{Measurement Types and Challenges}
Studies assessing LLMs' cognitive abilities typically rely on two types of indicators: \textbf{explicit outputs}, such as self-report responses of the model \citep{dasgupta2022language, hagendorff2023human} and counterfactual prompting \citep{roberts2024large}, or \textbf{implicit signals}, such as token probabilities \citep{lee-etal-2024-psycholinguistic, ullman2023large} and attention scores \citep{bazhukov-etal-2024-models}.
For example, \citet{roberts2024large} investigated the Fan Effect~\cite{anderson1974retrieval}, in which people take longer to retrieve information from memory when there is a large number of facts linked to that concept, in LLMs by measuring the token probability as a proxy for memory retrieval difficulty. 
Most studies employ a single measure, but it is ideal to collect several measures to establish convergent evidence of a given cognitive effect in LLMs. 

Translating human-designed experiments to LLMs poses several challenges. A key issue is that LLM responses are highly sensitive to prompt contents. Some studies address this by introducing prompt perturbation methods \citep{coda2023inducing, bazhukov-etal-2024-models} to account for variability in model outputs, and \citet{roberts2024using} propose injecting variability by perturbing model weights through the addition of dropout layers. Rather than adapting datasets from human studies, several works instead assess LLM cognition using existing NLP task datasets \citep{ying-etal-2024-intuitive, lu-etal-2024-emergent, shah-etal-2024-development} or curating new datasets tailored to specific research questions \citep{jones2022capturing, lal-etal-2024-cat, lv-etal-2024-coggpt, joshi-etal-2024-llms, jung-etal-2024-perceptions, lee-lim-2024-language}.
This approach allows researchers to tailor materials to LLMs, but sacrifices the high quality offered by datasets designed by social scientists. 
This approach also introduces challenges in comparing LLMs to humans due to a lack of pre-existing human data. Some studies collect human performance data for comparison \citep{lee-lim-2024-language}, but many omit this step \citep{ying-etal-2024-intuitive, joshi-etal-2024-llms}.
High quality human comparisons are important to probe details of LLM cognitive processes, since LLMs may achieve similar results to humans through fundamentally different cognitive mechanisms \cite{ivanova2025evaluate}. 



\section{Preliminary on Aspect and Humans}\label{sec:background}
Aspect is a particularly subtle meaning domain that is especially challenging for learners of English as a second language \cite{andersen1996primacy, bardovi2000tense}. The events in this study belong to the lexical aspect category ``Accomplishment'' and therefore come with an initial state, transition point, and final state. Accomplishment events prototypically pair up with \textit{perfective} grammatical aspect because the perfective also has an endpoint \citep{salaberry2024thirty}. Thus, it takes an advanced command of the language to combine the Accomplishment events in this study with an imperfective aspect that focuses on the initial state rather than the final state: such combinations are ultimately tied to \textbf{pragmatic text-level decisions} by speakers \citep{hopper1982aspect}. For instance, the non-prototypical coupling of imperfective aspect with Accomplishment events can signal to readers that something unusual may happen \citep{hopper1982aspect}. Thus, for native speakers or non-natives with advanced proficiency, aspect significantly influences narrative comprehension, shaping how readers process ongoing narratives and form situation models \cite{zwaan1995construction}, as well as how they incorporate events into their global discourse model \cite{mozuraitis2013younger}.

Aspect also plays a critical role in the comprehension of \textbf{causal relations} within narratives. Specifically, events marked by the \textit{imperfective aspect} are more likely to be interpreted as \textit{potential causes} for subsequent events. 
Consider the example in Table~\ref{tab:narrative-structure}:  When imperfective aspect is used (``Rob \textit{was washing} the dishes''), the action remains ongoing, making it a plausible cause of the subsequent loud noise. 
When perfective aspect is used (``Rob \textit{washed} the dishes''), the action is interpreted as completed, reducing its potential causal relevance to the unfolding narrative.

\begin{table}
\small
\centering
\begin{tabular}{|>{\columncolor[RGB]{226,237,143}} p{0.03\linewidth} |>{\columncolor[RGB]{204,229,255}} p{0.8\linewidth} |}
\hline
\multicolumn{1}{|>{\columncolor[RGB]{250,250,250}} p{0.05\linewidth}}{\vfill \rotatebox{90}{Intro} \vfill} 
& \vfill Rob and Alisha had a nice system going.\vfill \\
\hline
\multicolumn{1}{|>{\columncolor[RGB]{250,250,250}} p{0.05\linewidth}}{\vfill \rotatebox{90}{Filler} \vfill} 
& \vfill Each day they split up their duties and rotated them. Today, Alisha took care of the living room, and Rob was down for kitchen. \vfill \\
\hline
\multicolumn{1}{|>{\columncolor[RGB]{226,237,143}} p{0.05\linewidth}}{\vfill \rotatebox{90}{$C_1$} \vfill} 
& \vfill Rob *was washing* the dishes.
\vfill \\
\hline
\multicolumn{1}{|>{\columncolor[RGB]{226,237,143}} p{0.05\linewidth}}{\vfill \rotatebox{90}{$C_2$} \vfill} 
& \vfill Alisha watered the plants and started arranging flowers in one of their special vases.
\vfill \\
\hline
\multicolumn{1}{|>{\columncolor[RGB]{250,250,250}} p{0.05\linewidth}}{\vfill \rotatebox{90}{Filler} \vfill} 
& \vfill She felt bad because it had been a terrible day for her. She overslept, missed her bus, and was reprimanded by her boss. In her rush she forgot her purse. And on top of everything her lunch date did not show.
\vfill \\
\hline
\multicolumn{1}{|>{\columncolor[RGB]{226,237,143}} p{0.05\linewidth}}{\vfill \rotatebox{90}{Effect} \vfill} 
& \vfill Suddenly there was a loud noise.
\vfill \\
\hline
\end{tabular}
\caption{Structure of narratives designed by linguists to probe the impact of aspect on causal inferencing. The end effect has two possible causes, $C_1$ and $C_2$. When $C_1$ is imperfective (rather than perfective), it has a higher potential connection to the final effect.
}
\end{table}\label{tab:narrative-structure}

\subsection{Linguistic Design of Narrative Stimuli}
The narratives used in this study are adapted from linguistic research on aspect comprehension in English \citep{schramm1998aspect}. They contain an unusual combination of imperfective aspect and Accomplishment, providing a good opportunity to test whether LLMs are able to handle temporality in a fashion that parallels the behavior of human native speakers. 
Our study employs the inherent temporal structure of Accomplishment events to examine whether LLMs exhibit aspect-driven inference and processing patterns when reading narratives. To this end, narratives follow this format (e.g., Table~\ref{tab:narrative-structure}):
\begin{enumerate}[noitemsep]
    \item Early in the narrative, a potential cause event, $C_1$ is presented. The event is comprised of a prototypical closed-ended two-state Accomplishment situation (e.g., “dishes unwashed” $\rightarrow$ “dishes washed”) \citep{klein1994time}. 
    \item A second potential cause, $C_2$, follows $C_1$.
    \item A surprise-effect event, which can be causally linked to either $C_1$ or $C_2$, ends the narrative.
\end{enumerate}


This design allows us to influence the causal connection of $C_1$ to the final surprise event solely by changing the \textit{aspect markings} of the verb in $C_1$. When $C_1$ is imperfective (\textit{was washing}), it remains conceptually active as the reader processes the subsequent text, making both $C_1$ and $C_2$ plausible causes of the effect. When $C_1$ is perfective (\textit{washed}), it is interpreted as completed and recedes from the reader's focus, making $C_2$ the more likely inferred cause.

\subsection{Genuine Cognitive Comprehension}
To compare the performance of LLMs with that of humans in the arena of cognitive processes involved in understanding \citep{apidianaki2024language, ohmer2024form}, we investigate how LLMs first cognitively process temporal meanings expressed by linguistic aspect in narratives in working memory and eventually understand them in episodic memory.
In humans, aspect information is encoded in working memory during reading. It influences the integration of subsequent events into the discourse model, and is modulated by world knowledge regardless of age \citep{mozuraitis2013younger}. When the critical sentence is in the imperfective (``Rob was washing the dishes''), it is stored “in focus” in episodic memory after five sentences and can later be reactivated in working memory, because there is no clear endpoint \citep{schramm2016processing}. This in turn implies continued relevance as a potential cause for later events (``Suddenly there was a loud noise''). By contrast, when the critical sentence is in the perfective (``Rob washed the dishes''), the event itself is less accessible in episodic memory and less likely to be reactivated in working memory, since it is presented as completed, making it an unlikely or even impossible cause of the later surprise event. Only the after-effect of the event is available \citep{hart2009effects}. Furthermore, to signal in-focus status requires indirect, pragmatic understanding of aspect in a narrative context. Open-ended imperfective aspect in narratives in the foreground is highly unusual \citep{hopper1982aspect}, because events are in the past and are prototypically narrated in the perfective aspect. This non-prototypicality should lead to greater attention and in turn activation in participants’ working memory, which can then be used for causal inferencing.  

After processing in working memory during reading is complete, we investigate the pragmatic information available in episodic memory in humans. We query the discourse model regarding the cause for the surprise effect at the end of a story ("What was the reason for the loud noise?"). When the cause is in the open-ended imperfective, it is predicted to be stored accordingly more frequently in episodic memory than in the perfective.

It is also important for an assessment of human versus LLM performance to compare humans' cognitive processes involved in processing sentence-level semantic as opposed to text-level pragmatic meaning with those of LLMs. To accomplish this, we shifted to a semantic measure of aspect awareness at the sentence level. Truth value judgments obtained from humans' episodic memories ascertain whether a causal event was closed off ("dishes washed" in the perfective) or open-ended ("dishes unwashed" in the imperfective). In the former case, it would be less available for further cognitive processing, such as causal inferencing, than in the latter case.

\section{Expert-in-the-loop Probing Pipeline}\label{sec:pipeline}
To systematically investigate LLM behavior in comparison to human cognition, we develop an Expert-in-the-Loop probing pipeline. This pipeline facilitates controlled behavioral experiments with LLMs by integrating experimental material input, automatic prompt perturbation, and response parsing for preliminary result evaluation. 
Our philosophy is that the pipeline should be applied iteratively, to gather converging evidence, with experts assessing the intermediate results between iterations.

\paragraph{Overview}
Our pipeline is designed to ensure robust and interpretable LLM behavioral evaluations. The process consists of three core stages:
\begin{itemize}[noitemsep]
    \item Prompting – Constructing LLM prompts \textit{faithful} to human experimental designs.
    \item Prompt Paraphrasing – Introducing controlled variations to ensure \textit{robustness}.
    \item Models and Inference – Running experiments across multiple LLMs for supporting \textit{model agnostic} behaviors.
\end{itemize}
Below, we describe each stage in detail.


\paragraph{Prompting. }
For each behavioral experiment, we develop a task-specific prompt based on the instructions used in a corresponding human study. While instructions are modified to optimize clarity for LLMs (e.g. converting ``circle your answer'' into ``respond with your answer,'' and applying structured formatting), we maintain fidelity to the original human task to enable valid comparisons between LLM and human responses. 
Each prompt explains the task in clear, structured language, with an example to illustrate how the task should be completed.
Our pipeline also enforces two distinct components in prompts: general instructions, which provides the task details and an example, and a data format, which specifies how instances should be formatted. The general prompt serves as an instruction to the model, while the data format organizes task-specific input representations. For example, a general prompt may state: \textit{“Read the stories and follow the general instructions…”}, while a corresponding data format could be structured as \textit{“Stories: <story>, Question 1: <question1>”}. 
Finally, prompts are also formatted to display only one stimulus at a time to avoid order effects. For further prompt details see Appendix~\ref{sec:appendix_prompts}.

\paragraph{Prompt Paraphrasing. }
Prompt characteristics can significantly influence model responses. We therefore introduce controlled perturbations to both the general instruction component and the data format component of the prompt. \citet{sclar2024quantifying} show that formatting influences model responses; we adapt their structured formatting protocol \textsc{FormatSpread} to introduce 9 additional data format variations that alter whitespace, casing, ordering, and punctuation. To alter the general instructions, we follow \citet{wahle-etal-2024-paraphrase} who show that paraphrasing can also affect model responses. We apply paraphrasing categories of (1) discourse-based syntax and structural changes and (2) semantic changes to rephrase the general instructions while retaining meaning to create three versions of the general instructions. In all, we have a total of 30 unique prompt variations (3 general prompt × 10 data format) per prompt. By evaluating responses across these 30 variations, we avoid findings that are artifacts of a single prompt phrasing.

\paragraph{Models and Inference. }
We conduct inference on seven LLMs. Six are open weight models: Gemma2 with 9B and 27B parameters \cite{team2024gemma}, Llama3.1 with 8B and 70B parameters \cite{dubey2024llama}, and Qwen2 with 7B and 72B parameters \cite{yang2024qwen2}. One proprietary model is GPT4-o \cite{achiam2023gpt}.

We use the Hugging Face library \citep{wolf-etal-2020-transformers} for generation with open-source models, applying 4-bit quantization for efficiency when working with models exceeding 9B parameters. 
We use the OpenAI API to query GPT-4o.
Default generation hyperparameters are used in all cases.
Responses are parsed based on the format specified in the prompt, such that answers can be automatically extracted for analysis.


\section{Experiments and Results}
We examine three probes that evaluate various aspects of the narratives described in Section~\ref{sec:background}: truth value judgments (\S\ref{sec:exp_semantic}), word completion tasks (\S\ref{sec:exp_implicit}), and open-ended causal questions (\S\ref{sec:exp_causal}). The experiments query the LLMs based on the two versions (perfective/imperfective event in Cause 1) of the 16 narrative described in ~\S\ref{sec:background}. All experiments are facilitated by our Expert-in-the-Loop pipeline, meaning prompts are perturbed such that we have 30 versions of each prompt. Finally, we perform statistical analyses on the LLM responses to evaluate our hypotheses. Analyses used multilevel model analysis through the R package lme4~\cite{lmer}. Details of these analyses can be found in Appendix~\ref{sec:stats}.

\subsection{Experiment 1a: Truth Value Judgments}\label{sec:exp_semantic}
To assess whether LLMs understand the semantics of perfective and imperfective aspect in narratives, we probe their ability to infer completion in the Accomplishment events described earlier. Grammatical aspect determines whether an Accomplishment's final state is explicitly expressed (perfective) or whether the event is left open-ended (imperfective). For instance, the imperfective statement ``John was walking home'' does not entail the resulting final state ``John at home.''

\begin{figure*}[ht!]
    \centering
    {\includegraphics[width=0.4\textwidth]{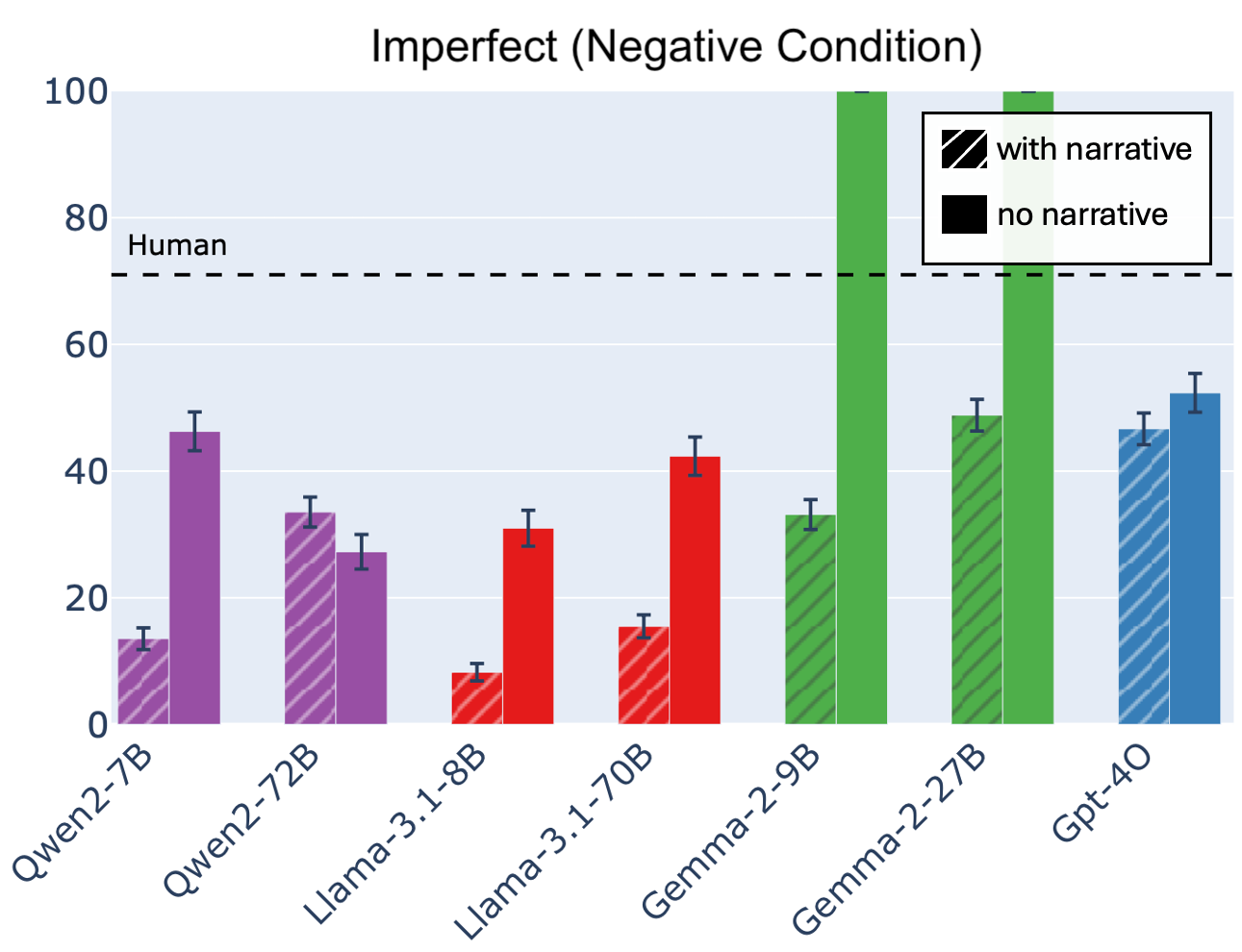}} 
    {\includegraphics[width=0.4\textwidth]{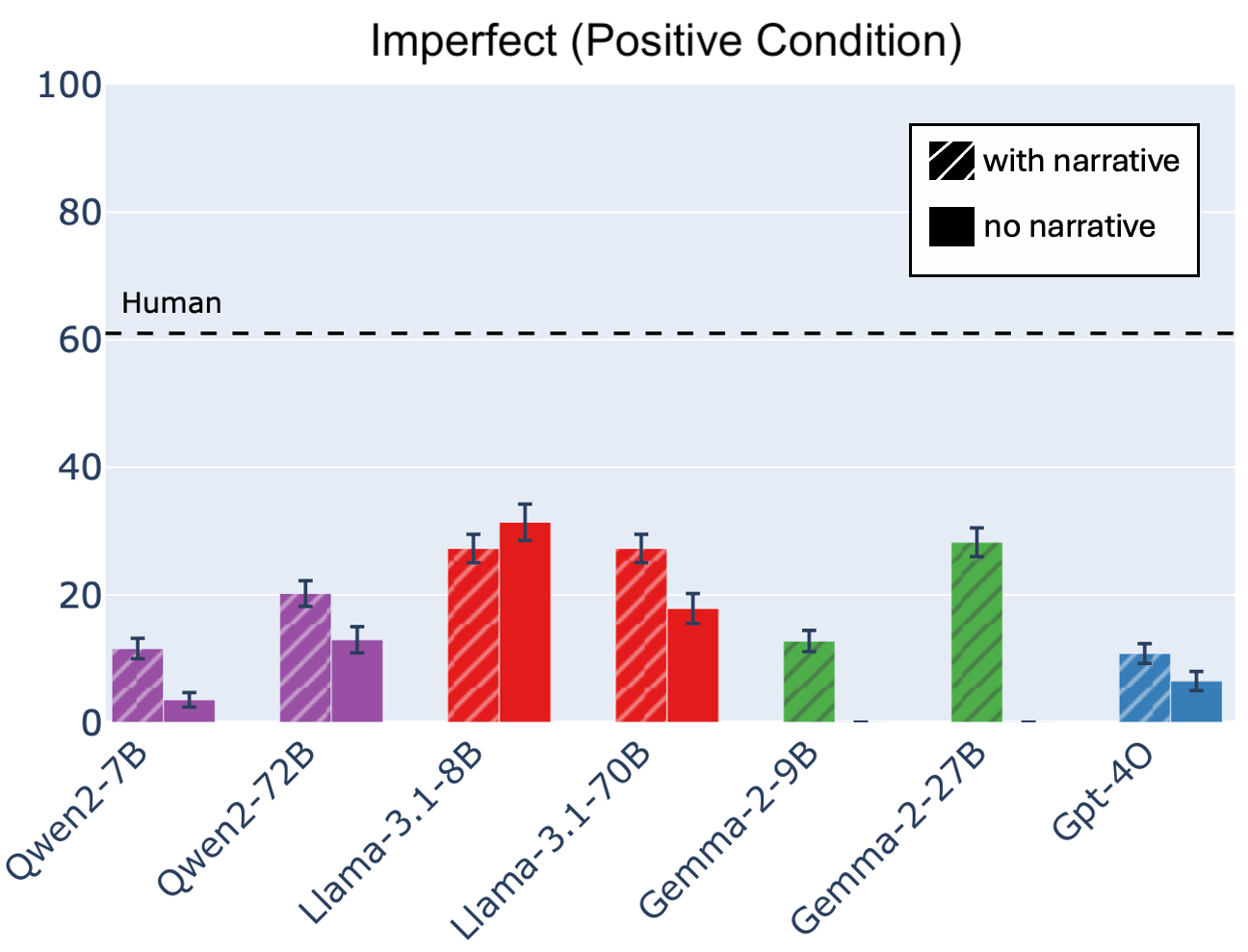}} 
    \caption{Accuracy in semantic truth-value judgments for events marked with imperfective aspect for LLMs, when the events are embedded within a narrative (shaded bars) versus not. For imperfective events, LLMs have much lower accuracy rates than humans when judging whether the event's resulting final state is valid. Further, LLMs seem to be heavily affected by the presence or absence of a narrative, especially when judging the negative polarity of final states. Notably, the presence or absence of a narrative changes responses in inconsistent directions. Error bars represent the standard error.}
    
    \label{fig:semantic_narrative}
\end{figure*}

This experiment tests whether LLMs correctly infer an event’s final state by evaluating truth-value judgments on the following two statements: (i) “John at home” (Positive inference), or (ii) “John not at home” (Negative inference).
Semantically, the positive final state is true in the perfective, and the negative final state is true in the imperfective.

\begin{tcolorbox}[colback=blue!5!white,colframe=blue!75!black,title=Semantics: Truth Value Judgments]
  \textbf{Story:} ...Lena ran down the stairs... 
  \tcblower
    \textbf{Truth Value Judgment:} Lena downstairs
\end{tcolorbox}

\paragraph{Setup.} LLMs are presented with narratives and two final state formulations (positive or negative) associated with the event in Cause 1. LLMs are instructed to ``\textit{respond with whether you think the phrase is `True' or `False' or `Both' or `Can't Decide' with respect to the story},'' where the phrase is the final state. These are adapted from the instructions in \citet{schramm2022implicit}.

\paragraph{Findings.}
We hypothesized that LLMs, given their strong syntactic and grammatical knowledge, would correctly identify the semantic implicature of grammatical aspect for Accomplishments' final states. 
Results (see Table~\ref{tab:tvj_narratives}) confirm that LLMs perform well for perfective events but struggle significantly with non-prototypical aspect (i.e., imperfective).
Although human truth-value judgments \citep{schramm2024intentional} also deviate some of the time from the semantically expected responses in non-prototypical imperfective aspect conditions within this narrative context, LLMs exhibit significantly lower accuracy (18\% compared to 71\% for humans). Notably, the truth-value judgment task is not forced;\footnote{``Both'' or ``Can't Decide'' are also options.} LLMs chose neither true nor false in 9\% of responses.  

\begin{table}[]
    \centering
    \small
    \begin{tabular}{l@{\hskip 1mm}|@{\hskip 1mm}c@{\hskip 1mm}|c@{\hskip 1mm}|@{\hskip 1mm}c@{\hskip 1mm}c}
         \textbf{Aspect} & \textbf{Polarity} & \textbf{Target} & \textbf{LLM} & \textbf{Human}  \\\hline
         Perfect & Positive & True & 88\% & 88\% \\ \hline
         Imperfect & Negative & True & \textbf{18\%} & 71\%\\\hline
         Perfect & Negative & False & 89\% & 93\% \\ \hline
         Imperfect & Positive & False &   \textbf{35\%} & 61\% \\ \hline
    \end{tabular}
    \caption{
    LLM semantic understanding of aspect. Narratives include an event in either the imperfective ("Lena was running downstairs") or perfective ("Lena ran downstairs") aspect. LLMs then evaluate the truth of the resulting state with positive or negative polarity ("Lena (not) downstairs"). The target column shows the semantically correct truth value. \textbf{Results show that LLMs perform significantly worse than humans in non-prototypical imperfective conditions.} Human data can be directly accessed in \citet{schramm2024intentional}.
    }
    \label{tab:tvj_narratives}
\end{table}

\subsubsection*{Experiment 1b: Influence of Narrative Context}
The narrative context in Experiment 1a may pose a challenge for LLMs since the Accomplishment-imperfective combination is rarely seen in narratives, which could interfere with the LLMs' ability to display its grammatical knowledge in its response. To address this possibility, we repeat the experiment with the narrative context omitted. However, with no narrative context performance in the imperfective condition remains low (see Figure~\ref{fig:semantic_narrative}), reinforcing that the difficulty is inherent to aspect processing and not an artifact of the unusual narrative structure. Interestingly, LLMs show both positive and negative shifts in accuracy across models when narratives are removed, suggesting variability in how different architectures comprehend aspect with and without context.

\paragraph{Findings.}
In summary, our experiments reveal that LLMs have significant sensitivity to narrative context when evaluating events in the imperfective aspect. Regardless of the presence of narrative context, \textbf{LLM responses are significantly less accurate for imperfective events} (compared to perfective). The perfective events in narratives are much more common, suggesting that LLMs may rely on high-frequency linguistic patterns more than a deep mastery of aspect semantics in their responses. 

This experiment evaluates self-reported interpretations from LLMs, which may not fully capture their understanding of aspect semantics and grammatical knowledge. However, it is notable that their responses in these experiments diverged sharply from humans' despite the variety of prompts presented in the experiments.


\subsection{Experiment 2: Word Completion Task}\label{sec:exp_implicit}
Do LLMs, like humans, treat aspect in narratives as a temporary signal to eventually retain or encode an event? Aspect in narratives influences temporary human working memory and permanent event encoding: Prior studies show that when an event is described with imperfective aspect, humans infer its ongoing relevance and are more likely to retain it in working memory \citep{mozuraitis2013younger}. To investigate whether LLMs exhibit a similar response, we conduct a word-completion experiment, a widely used measure of concept activation in working memory \cite{tiggemann2004word} that has also been used in linguistic aspect studies \citep{schramm2016processing}.

\begin{tcolorbox}[colback=blue!5!white,colframe=blue!75!black,title=Word Completion Task]
  \textbf{Story:}  ... she *was eating* the apple ... 
  \tcblower
    \textbf{Complete the word:} A P \_ \_ \_
\end{tcolorbox}

If an event remains active in working memory, its associated key noun is expected to be retrieved more frequently. The frequency of specific word completions serves as an indicator of \textbf{concept accessibility}, reflecting how strongly a word is activated by underlying cognitive processes. Unlike other measures of word activation, such as timed lexical decision tasks, word completion is a feasible assessment tool for LLMs. This task is grounded in the priming effect, a phenomenon in human cognition where exposure to a stimulus influences subsequent recall. Prior research suggests that LLMs exhibit similar priming-based activation patterns, making word completion a viable proxy for assessing concept accessibility in LLMs \cite{roberts2024large, jumelet2024language}.\footnote{However, because LLM attention mechanisms differ fundamentally from working memory in humans, LLM responses to this task should be interpreted with caution.}
\paragraph{Setup.} In addition to a narrative, LLMs are presented with two partial words (e.g., A P \_ \_ \_) and asked to fill in the blanks to create a complete word (e.g., A P P L E). One partial word is a distractor, and the other can be completed with a word from the event in the critical Cause 1 section of the narrative. Crucially, the \textit{location} of the partial word question is also varied to test the extent to which the cognitive processes in LLMs are similar to those in humans' working memory. In the ``Near Cause 1'' condition, the word completion probe is located after the sentence with the target word, allowing us to estimate the extent to which the word is activated because the imperfective is nonprototypical at the Cause 1, thus putting the event in focus. In the ``Near Effect'' condition, the word completion probe is located after the effect, and we investigate the extent to which the target word is reactivated by the surprise effect at the end of the narrative, as would be expected when Cause 1 has the unbounded imperfective aspect. 

\begin{figure}[t!]
    \centering\vspace{-7mm}
    \includegraphics[width=0.8\linewidth, trim={0 0 0 0},clip]{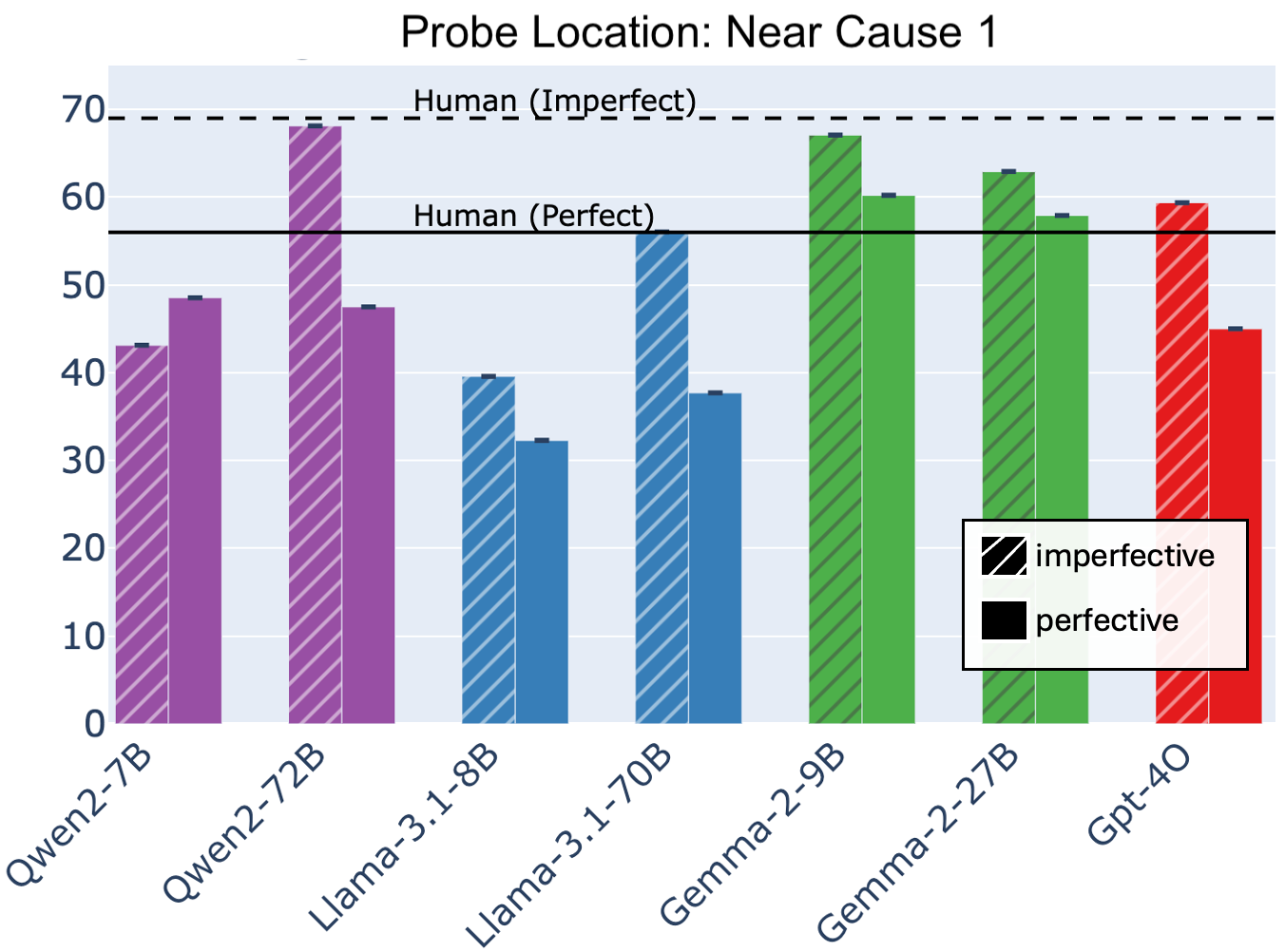}
    \includegraphics[width=0.8\linewidth, trim={0 0 0 0cm},clip]{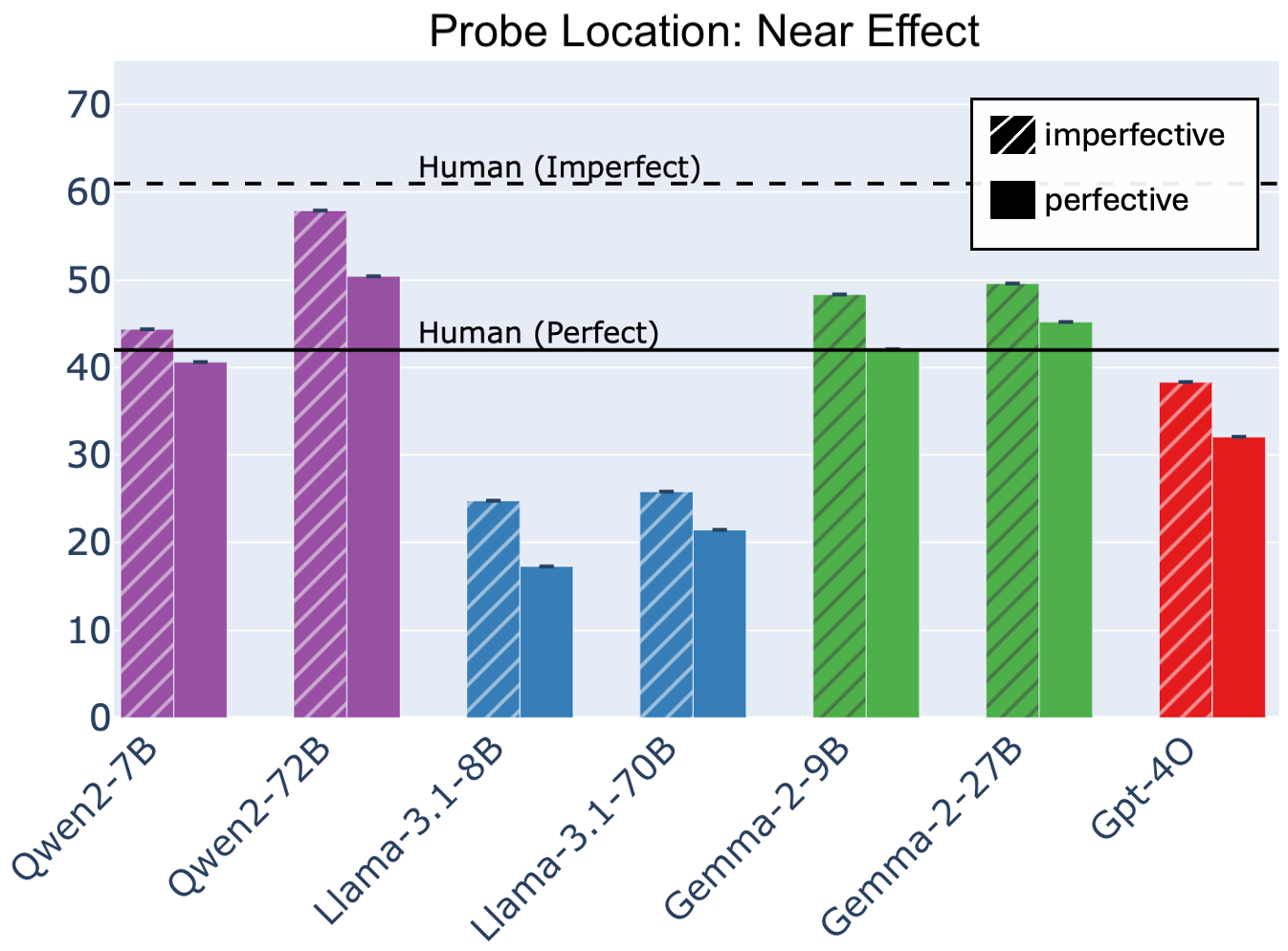}
    \vspace{-4mm}
    \caption{Frequencies at which word completion rates match the target word from Cause 1 across models. Shaded bars are for imperfective aspect in Cause 1. LLM completions have significantly higher match frequencies when the probe directly follows Cause 1 (top) and are reduced after the effect (bottom).}
    \label{fig:wc}
\end{figure}

\paragraph{Findings.}
Our results show that LLM responses roughly align with human patterns when the word completion occurs near Cause 1. However, the frequency of responses that point to the target word decrease (on average 33\%) when the completion is placed near the Effect, and differences between imperfective and perfective shrink (Figure~\ref{fig:wc}). This suggests that while LLMs may in the short term attend more to pragmatic nonprototypicality information regarding imperfective narrative, they lack \textbf{distal causal narrative integration capabilities} akin to human memory mechanisms.
Additionally, because imperfective aspect in the earlier foregrounded events is highly non-prototypical, LLMs may be attending to these events due to their statistical rarity, rather than demonstrating an implicit understanding of their pragmatic implications.

\subsection{Experiment 3: Open-ended Causal Questioning}\label{sec:exp_causal}

As discussed in Section~\ref{sec:background}, imperfective aspect in narratives carries causal implications, influencing how humans interpret event relationships in episodic memory. In this experiment, we examine whether LLMs’ causal inferences are similarly affected by aspect. Recall the narrative structure described in \S~\ref{sec:background}, in which a surprising effect is introduced, and there are two potential causes:
When Cause 1 is presented in the imperfective aspect, it is more likely to be perceived as ongoing, consequently increasing its likelihood of being inferred as the cause.
When Cause 1 is presented in the perfective aspect, it is more likely to be perceived as completed, reducing its consideration during causal inference.

\begin{tcolorbox}[colback=blue!5!white,colframe=blue!75!black,title=Open-ended Causal Questioning]
  \textbf{Story:}  ... Suddenly there was a loud noise.
  \tcblower
    \textbf{Question:} Why was there a loud noise?
\end{tcolorbox}

Human studies confirm that imperfective aspect significantly increases the likelihood of Cause 1 being inferred as the event's cause \citep{schramm2016processing}. We assess whether LLMs exhibit similar causal reasoning patterns.

\paragraph{Setup.}
An open-ended causal question follows the narrative, asking the LLM to infer the most plausible cause of the final effect. We measure how frequently Cause 1 and Cause 2 are identified as causing the effect. 
Since manual annotation of the open-ended LLM responses is impractical at scale, we employ OpenAI’s API to assist with response classification (see Appendix~\ref{sec:appendix_prompts} for details). To ensure reliability, one author manually coded 128 responses for comparison. We measured inter-annotator agreement between the manual and automated annotations using Cohen’s Kappa and found $\kappa = .93$, indicating strong reliability.

\begin{figure}
    \centering
    \includegraphics[width=\linewidth,trim={0.1cm 0 1.2cm 0},clip]{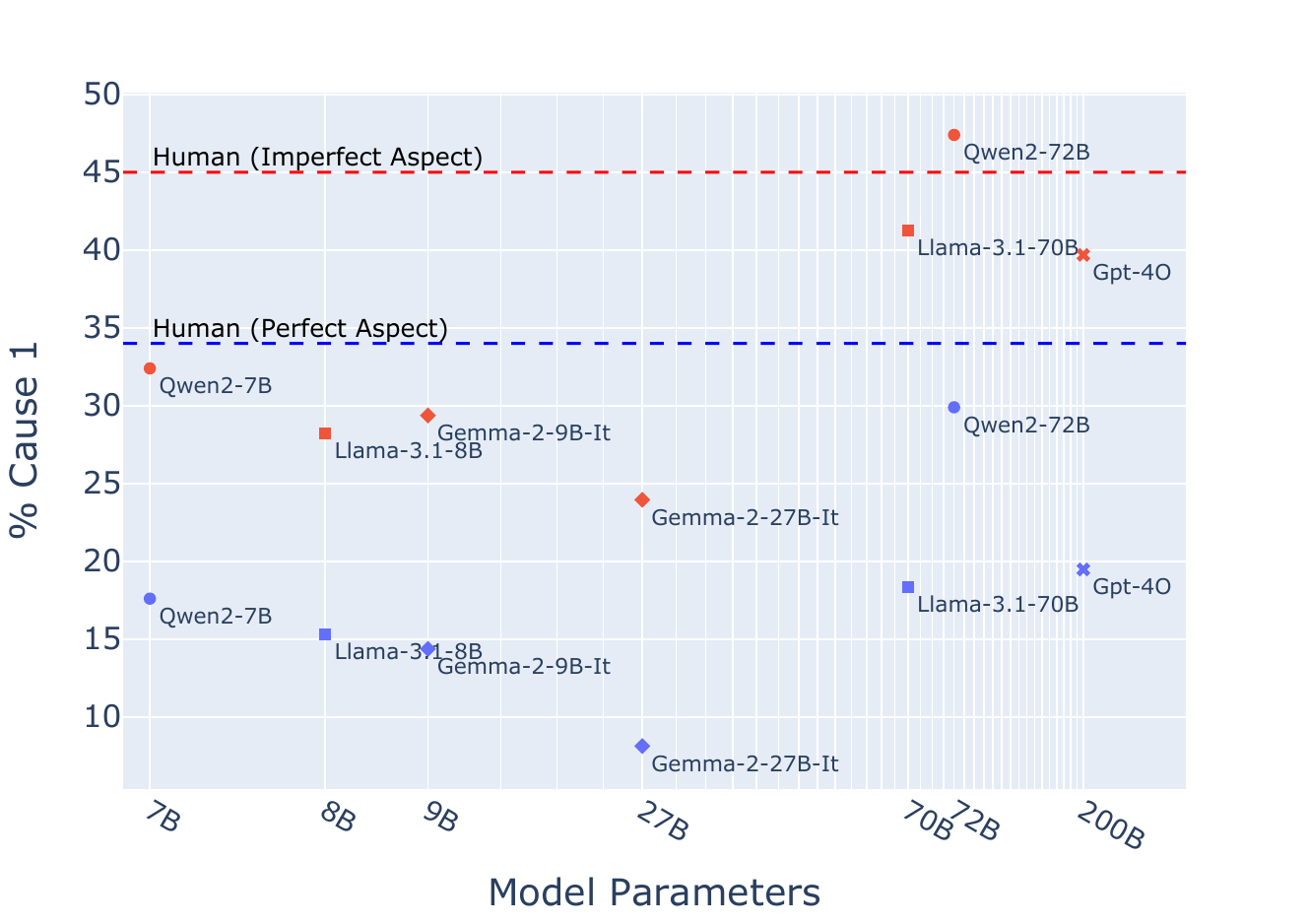}
    \caption{As LLM parameter size increases, there is a trend towards more human-like causal inferences with respect to the Cause 1 event when Cause 1 is in the imperfective. When Cause 1 is in the perfective, LLMs are consistently below human causal inference rates.
    }
    \label{fig:causality}
\end{figure}

\paragraph{Findings.}
The results of LLM causal inferences are presented in Figure~\ref{fig:causality} and compared to previous human results \citep{schramm2016processing}. We observe that LLMs, like humans, are also more likely to infer that Cause 1 caused the effect in the imperfective condition (Fig~\ref{fig:causality}). However, most LLMs make this inference substantially less frequently than humans do.

The LLMs that score closest to humans on the open-ended causal inference question are notably not necessarily the highest scores in our other aspect-related experiments (e.g., Llama-3.1-70B does well in this task, but was not close to humans in Experiments 1a, 1b, and 2). Qwen2-72B also has a strong performance, and notably also did very well in the word completion task in the critical ``Near Effect'' condition, suggesting that the word completion task may be a valuable probe for LLMs.  Finally, all LLMs tested were much less likely than humans to offer a \textit{perfective} Cause 1 as their answer. This may again point to an over-reliance on prototypicality, or perhaps to a lack of distal causal capabilities -- LLMs tend to lose attention on Cause 1 by the end of the narrative, as evidenced by the ``Near Effect'' condition results in Experiment 2.

\section{Discussion}
Our findings indicate that LLMs process aspect in narratives fundamentally differently from humans. While LLM truth value judgments in Experiment 1 mostly align with humans in prototypical cases, they deviate sharply in non-prototypical contexts. 
Interestingly, despite the poor truth-value judgments in non-prototypical cases, LLMs can readily articulate the definition of aspect in academic discussions.
We speculate that aspect is one of several areas in which LLMs excel in declarative knowledge (e.g., reciting a definition) while failing in implicit application (e.g., making truth-value judgments following the definition). This contrasts sharply with humans, who may lack declarative knowledge of aspect but consistently demonstrate \textbf{implicit understanding} in real-world judgments. Future research should explore how the tension between declarative and implicit knowledge extends to other linguistic and cognitive domains.

In Experiment 2, we observe that while LLMs seem to detect pragmatic non-prototypicalities within immediate contexts (``near Cause 1''), \textbf{LLMs tend to not maintain causal pragmatic focus on imperfective events} over the distance of the narrative (``near Effect''). If LLMs constructed temporary internal situation models as humans do, we would expect a more maintained focus. This raises broader questions about how LLMs track narrative elements, such as point of view, and temporal progression.

Experiment 3 also explores whether LLMs grasp the causal implications of aspectual distinctions by probing the contents of the narrative situation model. Our results suggest that larger models exhibit more human-like, albeit less robust, causal reasoning, even though in previous experiments larger models showed no clear improvements in aspect-based temporary pragmatic focus or semantic aspect comprehension. However, we see that while a substantial portion of humans (one-third) ascribe the event in Cause 1 to the effect even when Cause 1 is in the perfective, LLMs are significantly less likely to do so. This suggests that humans approach narrative interpretation with \textbf{more flexibility}, striving for coherence and constructing internal situation models in ways that LLMs do not.

\paragraph{Pragmatic Understanding.} Overall linguistically, the performance of LLMs appears to be due to a lack of pragmatic context-level understanding \citep{beuls2024humans}. The semantic exploration of truth value conditions connected with prototypical and non-prototypical pairings of Accomplishment events possessing final states with bounded perfective aspect and unbounded imperfective aspect suggests that LLMs represent aspect distributionally, rather than based on the concepts expressed by aspect, similar to learners of English \citep{salaberry2024thirty}. The accuracy rates for LLMs are high in the bounded perfective when prototypically paired with bounded Accomplishment events with a final state, and they are low when the events are non-prototypically combined with unbounded imperfective aspect.

It is thus not surprising that LLMs also appeared to differ from humans when we compare their respective cognitive processes involved in the understanding of (non)prototypicality and causality, echoing recent findings (e.g., \citet{apidianaki2024language} and \citet{ohmer2024form}). Although probes in close proximity to Cause 1 have accuracy rates in the right direction, the rates are lower for probes in close proximity to the final effect, and comprehension of aspectual meanings and the distal inference of causality largely failed. In the eventual cognitive situation model, shortcomings were also apparent. LLMs' pragmatic understanding is generally lower than humans', and the lack of a connection between cause and effect in the perfective aspect is exaggerated. This may be due to the distributional (rather than meaning-based) prototypical aspect understanding found on the semantic level. Speculatively, the near-human understanding of the imperfective in the situation model may be a ``compensatory over-emphasis'' to the non-prototypicality of this aspect rather than human-like behavior, since the semantic accuracy was fairly low. 

\paragraph{Effects of Model Type and Size.} 
We compare three state-of-the-art model families (Gemma, Llama, Qwen) and one proprietary model (GPT-4o), evaluating both small and large model sizes to examine differences in family architecture and parameter scaling. While model families show statistically significant differences across tasks, no model or model family demonstrates consistent mastery or failure in aspect-related processing. This suggests that LLMs do not process aspect using human-like mechanisms. Moreover, there is no clear relationship between model performance on structured aspectual tasks and their performance on open-ended causal inference tasks, indicating that \textbf{aspect comprehension and causal reasoning may be processed separately in LLMs}. Interestingly, we only observe a clear distinction between larger and smaller models in the open-ended causal inference experiment: larger models demonstrate a stronger grasp of imperfective causal implications. It remains uncertain whether further scaling will bridge the gap between LLMs and human-like aspectual processing.

\paragraph{Conclusion} Through a series of targeted experiments, we assess LLMs’ cognitive capabilities in processing linguistic aspect. Our findings indicate that LLMs over-rely on prototypicality, exhibit inconsistent aspectual judgments, and often fall short of performing human-like aspect-mediated causal inference, raising concerns about their ability to fully comprehend narratives. 

\section{Limitations}

This work relies on ``self''-report from large language models in response to prompts designed to probe specific functionalities. The epistemology of LLM responses, i.e., how well LLM self-report reflects internal states, is unclear. We make efforts to mitigate this risk by conducting multiple experiments, paraphrasing our prompts, and measuring multiple signals from the responses. 

This work also evaluates LLMs on narrative comprehension in controlled experimental states and without interaction. It is always possible that with more dialogue, or with different prompt engineering, results would change. However, if LLM comprehension of a topic is robust, it should not demand elaborate prompt engineering. With this in mind, we believe our results are informative, although they should be contextualized within the specific parameters of our experiments.

\section{Ethics}

We note that the theoretical framework for LLM probing discussed in this work, if misused or misapplied, may lead to incorrect conclusions about the nature of LLMs and LLM cognition. This can be misleading and in rare cases causes personal upset and concern \cite{nprgoogleeng}. We therefore maintain that the framework used in this work is best applied responsibly by researchers working in interdisciplinary teams.


\section*{Acknowledgments}
This research was supported in part by the Doctoral Dissertation Fellowship from the University of Minnesota for the first author.

\bibliography{custom}

\appendix
\section{Appendix}
\subsection{Statistical Analyses}\label{sec:stats}
We conduct statistical analyses on experimental results to assess the significance of the experimental manipulations, and to control for random effects inherent in the individual narratives. In our analyses we set $\alpha = .01$. We use the lme4 package~\cite{lmer} in R to perform analyses and the emmeans package \cite{emmeans} to calculate the estimated marginal means. F values are calculated using Satterwaithe's method \cite{luke2017evaluating}.

\subsubsection{Experiment 1: Truth Value Judgments}
This experiment has two independent variables, aspect and polarity of final state (positive or negative). To model LLM responses, we define aspect and polarity as fixed effects, and narrative type as a random effect. LLM responses are coded to binary variable indicating correctness. Our model is:

\texttt{accuracy} $\sim$ \texttt{aspect * polarity + (1 | narrative)}

We find a significant main effect of aspect, $F = 66.5, p < .01$, in which imperfective has lower LLM accuracy than perfective ($M = .41, SE = .02$ for imperfective, $M = .46, SE=.02$ for perfective). There is also a significant main effect of polarity, $F = 10363, p < .01$ with performance being much higher in the positive than in the negative condition ($M = .76, SE = .02$ for positive and $M = .12, SE = .02$ for negative) and a significant aspect and polarity interaction. The interaction between the two is also significant at $F = 661.5, p < .01$. All pairwise comparison between aspect and polarity condition combinations were significant (Bonferonni corrections for multiple comparisons were applied).

\subsubsection{Experiment 2}
This experiment has two independent variables, aspect (perfective or imperfective) and the probe location in the prompt (early or late). To model LLM responses, we define aspect and location as fixed effects, and narrative type as a random effect. LLM responses are coded to binary variable indicating presence or absence of the target word in the response. The model is:

\texttt{response} $\sim$ \texttt{aspect * location + (1 | narrative)}

We find a significant main effect of aspect, $F = 92.4, p < .01$, in which imperfective has lower accuracy than perfective ($M = .41, SE = .04$ for imperfective, $M = .49, SE =.04$ for perfective). There is also a significant main effect of probe location, $F = 282.2, p < .01$ with responses happening much more frequently in the early than in the late condition ($M = .52, SE = .04$ for early and $M = .39, SE = .04$ for late). The interaction between the two approaches significance $F = 5.9, p = .02$. All pairwise comparison between aspect and probe location condition combinations were significant (Bonferonni corrections for multiple comparisons were applied).

\subsubsection{Experiment 3}
This experiment has two independent variables, aspect and probe location. We define the same model as in Experiment 2.

We find a significant main effect of aspect, $F = 98.5, p < .01$, in which imperfective has lower accuracy than perfective ($M = .35, SE = .06$ for imperfective, $M = .17, SE =.06$ for perfective). Although we have no hypothesis for probe location, there is a significant main effect of probe location, $F = 25.3, p < .01$ with responses happening much more frequently in the late condition than in the early condition ($M = .22, SE = .30$ for early and $M = .39, SE = .04$ for late). The interaction between the two approaches significance at $F = 2.4, p = .02$.

\subsection{Web Application}\label{sec:appendix_web}
\begin{figure*}
    \centering
    \includegraphics[width=\linewidth]{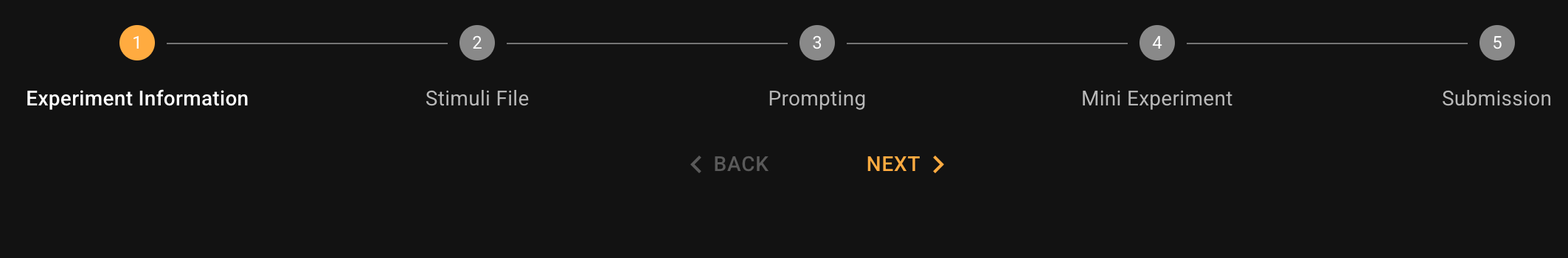}
    \vspace{.3em}
    \includegraphics[width=\linewidth]{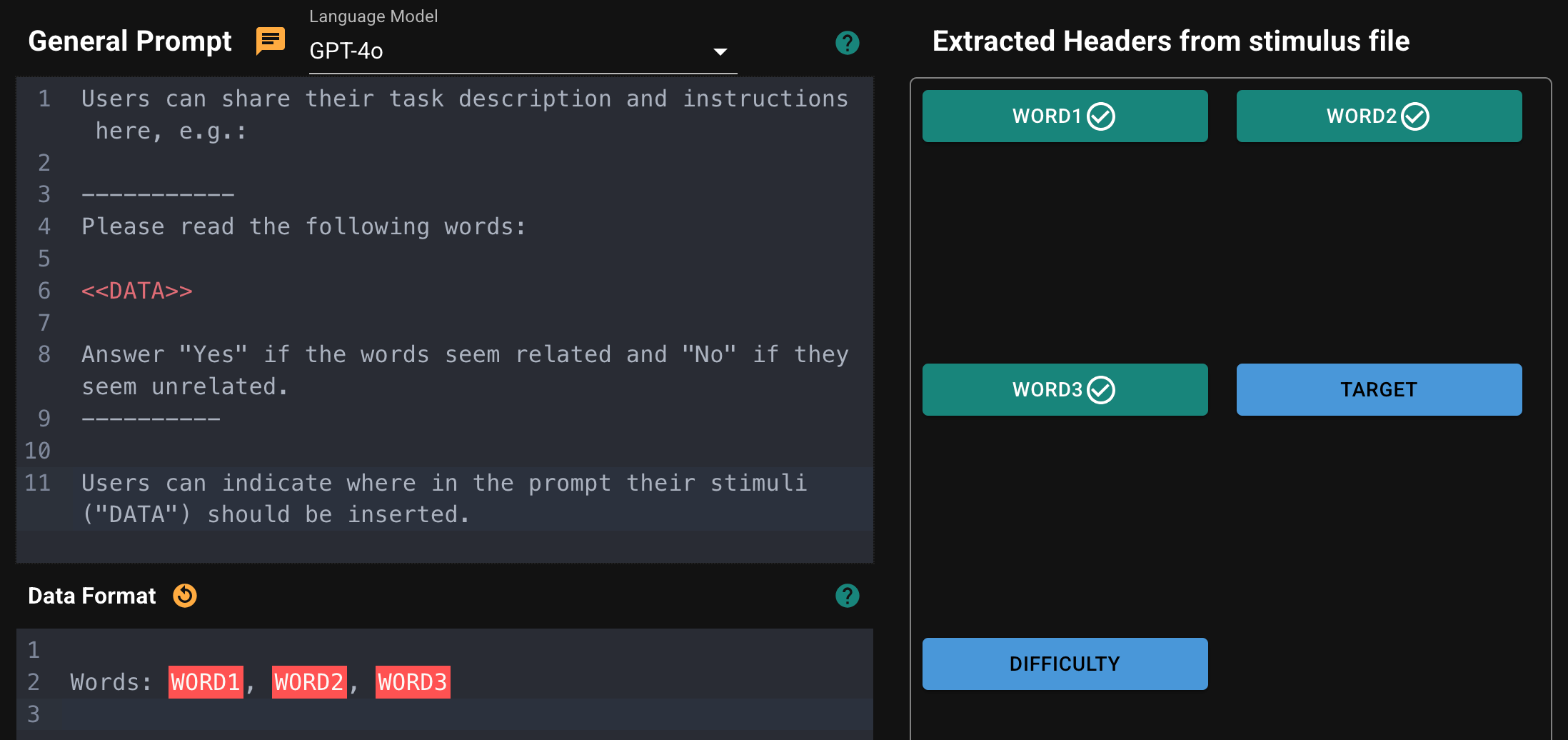}
    \vspace{.3em}
    \includegraphics[width=\linewidth]{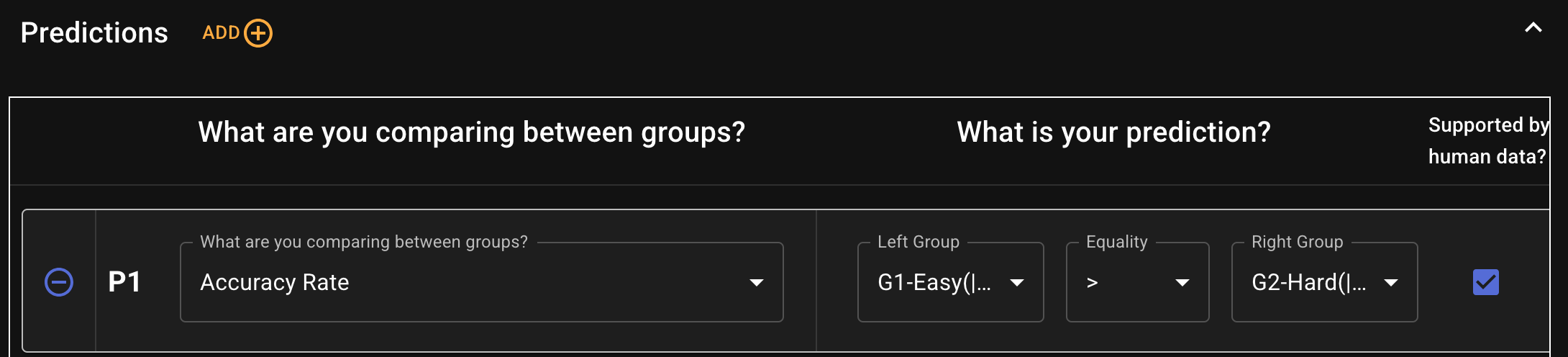}
    \caption{Important components of our web application for collaborating with cognitive scientists. \textbf{Top}: A navigation bar showing the pipeline that cognitive scientists complete to submit an experiment to the pipeline. \textbf{Middle}: Cognitive scientists can share task instructions and select which columns from their uploaded stimulus file should be included in the prompt (we then post-process the prompts and create 30 paraphrased versions). \textbf{Bottom}: Cognitive scientists identify independent variables from their uploaded stimulus file and define groups based on these independent variables.}
    \label{fig:web-interface}
\end{figure*}

An overview of the highlights of our web application for collaborating with cognitive scientists is shown in Figure~\ref{fig:web-interface}. Further details can be found in ~\citet{de2025framework}. Our collaboration pipeline is generalizable, allowing users to upload generic stimulus files that are the basis of the remainder of the pipeline. Users can select which parts of the stimulus files should be shown to the LLM during experiments and specify the independent variables and resulting groups that should be analyzed.

\subsection{Prompts}\label{sec:appendix_prompts}
The original prompts (i.e., before paraphrasing) are included for reference. The prompts for Experiment 1 are in Tables~\ref{tab:tvjnarrative} and ~\ref{tab:tvjsemantic}. The prompt for Experiments 2 and 3, word completion and causal question inference tasks, is in Table~\ref{tab:prompt1}.

\begin{table*}[h]
  \renewcommand{\arraystretch}{1.2} 
  \small
  \begin{tabular}{p{0.9\linewidth}} 
    \texttt{\#\#\# General Instructions:} \\ 

    \texttt{This study evaluates your ability to handle two simultaneous tasks:} \\

    \texttt{\hspace*{2em}- Task 1: Read and understand stories well enough to answer comprehension questions correctly.} \\
    \texttt{\hspace*{2em}- Task 2: Occasionally complete "word edges" as quickly as possible.} \vspace{0.5em} \\

    \texttt{\#\#\# What is a "word edge"?} \\ 

    \texttt{- A word edge consists of one or two starting letters of a word, followed by blanks. Your task is to fill in the blanks to form a valid English word.} \\
    \texttt{- Complete the word with the first valid word that comes to mind. The word can be a compound word (e.g., SETUP or LUNCHBOX) or include endings (e.g., STICKS or BUSHES).} \\
    \texttt{- Important:} \\
    \texttt{\hspace*{2em}- Fill all blanks (no extra letters or fewer letters than the blanks provided).} \\
    \texttt{\hspace*{2em}- Do not use personal names.} \vspace{0.5em} \\

    \texttt{\#\#\# Example of a Word Edge:} \\ 

    \texttt{- BR \_ \_ \_ \_ \_} \\
    \texttt{\hspace*{2em}- Correct answers:} \\
    \texttt{\hspace*{4em}- BRACKET} \\
    \texttt{\hspace*{4em}- BREAKIN} \\
    \texttt{\hspace*{4em}- BRUSHES} \\
    \texttt{\hspace*{2em}- Incorrect answers:} \\
    \texttt{\hspace*{4em}- BREAKAGE (too many letters)} \\
    \texttt{\hspace*{4em}- BRIDG\_ (too few letters)} \\
    \texttt{\hspace*{4em}- BRIDGIT (personal name)} \vspace{0.5em} \\

    \texttt{\#\#\# Instructions for the task:} \\ 

    \texttt{1. Read the stories provided.} \\
    \texttt{2. Occasionally, you will encounter two partial word edges.} \\
    \texttt{3. Some stories are split into parts, and the word edges may appear in the middle.} \\
    \texttt{4. Other stories are shown in full, with word edges at the end.} \\
    \texttt{5. Fill in the blanks for the first partial word that comes to mind.} \\
    \texttt{6. Then, complete the second word edge.} \\
    \texttt{7. Continue reading the story.} \\
    \texttt{8. After each story, you will be asked a question to test your memory of its content.} \vspace{0.5em} \\

    \texttt{\#\#\# Answer Format:} \\ 

    \texttt{- Question 1:} \\
    \texttt{\{WORD1\}} \\
    \texttt{\{WORD2\}} \\
    \texttt{- Question 1:} \\
    \texttt{\{ANSWER\}} \vspace{0.5em} \\

    \texttt{Note: Do not add any explanations or repeat stories/questions. Strictly follow the Answer Format provided.} \vspace{0.5em} \\

    \texttt{Story Part 1: \{\{STORY PART 1\}\}} \vspace{0.5em} \\

    \texttt{Question 1:} \\
    \texttt{\{\{QUESTION 1\}\}} \vspace{0.5em} \\

    \texttt{Story Part 2: \{\{STORY PART 2\}\}} \vspace{0.5em} \\

    \texttt{Question 2: \{\{QUESTION 2\}\}} \\
  \end{tabular}
  \caption{Original prompt (prior to prompt perturbation) used for word completion and causal inference experiments.}\label{tab:prompt1}
\end{table*}

\paragraph{Evaluating open-ended causal question responses}
The prompt used in the automatic scoring of the LLM responses to the causal inference questions in Experiment 3 is shown in Table~\ref{tab:causalprompt}.

\begin{table*}[h]
  \renewcommand{\arraystretch}{1.2} 
  \small
  \begin{tabular}{p{0.9\linewidth}} 
    \texttt{You are evaluating the answer to a question about a story. You will see the story, an extracted sentence from the story, and the question and answer.} \vspace{0.5em} \\ 

    \texttt{The extracted sentence will either be imperfect or perfect tense. We call this grammatical aspect, and it can be interpreted to have temporal meaning for the story (the aspectual meaning). The question will be about what *caused* an outcome at the end of the story.} \vspace{0.5em} \\

    \texttt{Your job is to decide whether the answer refers to information from the extracted sentence (true/false). If the answer refers to the aspectual temporal meanings of the extracted sentence's verb or to its non-completion, respond with True. If the answer does not refer to information from the extracted sentence or indicated that that the verb has been completed, respond with False.} \vspace{0.5em} \\

    \texttt{Think only within the given materials, and regardless of whether you agree or disagree with the answer, determine whether the answer refers to information from the extracted sentence. Provide a True or False response and explain your rationale clearly. Do not evaluate whether the answerer is right or wrong. Accept the answer as valid, and if the answer refers to the aspectual meanings of the extracted sentence's verb or to its non-completion, respond with True. If the answer does not refer to information from the extracted sentence or indicated that that the verb has been completed, respond with False.} \vspace{0.5em} \\

    \texttt{[Story]} \\
    \texttt{\{\{full\_story\}\}} \vspace{0.5em} \\

    \texttt{[Extracted Sentence]} \\
    \texttt{\{\{extracted\_sentence\}\}} \vspace{0.5em} \\

    \texttt{[Question]} \\
    \texttt{\{\{question\}\}} \vspace{0.5em} \\

    \texttt{[Answer]} \\
    \texttt{\{\{answer\}\}} \vspace{0.5em} \\

    \texttt{[ExtractedConsideredCause (true/false)]:} \\
    \texttt{rationale:} \\

  \end{tabular}
  \caption{Prompt that was used to automatically score responses to all LLM causal question responses via the OpenAI API.}\label{tab:causalprompt}
\end{table*}

\begin{table*}[h]
  \renewcommand{\arraystretch}{1.2} 
  \small
  \begin{tabular}{p{0.9\linewidth}} 
    \texttt{This is a test designed to test your ability to manage two different tasks at once: Task 1: to read and comprehend about a story well enough to evaluate a phrase that may refer to the story, and Task 2: to assess how accurate your evaluation of the phrase is. The phrases are 2-5 words long and are not complete sentences. You should decide *whether the phrase is true* with respect to the story. You should respond with "True" or "False" or "Both" or "Can't Decide" according to the first understanding that comes to mind that fits the story. Specifically, you'll first read a story. When you get to the end of a story, the last sentence will repeat. Below this last sentence is a phrase that may refer back to the story. Immediately respond with whether you think the phrase is "True" or "False" or "Both" or "Can't Decide" *with respect to the story.* Make sure your response contains ONLY your truth value judgment (True/False/Both/Can't Decide). Here's an example:} \vspace{0.5em} \\ 

    \texttt{STORY: Nancy and Chris were moving into a new apartment. They decided that Nancy would get the big bedroom and Chris would get the garage space. Nancy and Chris decided to move their furniture into their rooms first. When Chris left with the truck to get the furniture, Nancy PAINTED her wall.} \vspace{0.5em} \\

    \texttt{LAST SENTENCE: Nancy painted her wall} \\
    \texttt{PHRASE: whole wall with fresh paint} \\
    \texttt{Is this phrase true with respect to the story?} \\
    \texttt{Option 1 - True} \\
    \texttt{Option 2 - False} \\
    \texttt{Option 3 - Both} \\
    \texttt{Option 4 - Can't Decide} \vspace{0.5em} \\

    \texttt{Response: True} \vspace{0.5em} \\

    \texttt{STORY: \{\{STORY TO INCLUDE IN PROMPT\}\}} \vspace{0.5em} \\

    \texttt{LAST SENTENCE: \{\{LAST SENTENCE\}\}} \\
    \texttt{PHRASE: \{\{PHRASE\}\}} \\
    \texttt{\{\{OPTIONS\}\}} \vspace{0.5em} \\

    \texttt{Response:} \vspace{0.5em} \\

  \end{tabular}
  \caption{Prompt that was used for Experiment 1a.}\label{tab:tvjnarrative}
\end{table*}

\paragraph{Experiment 1: Truth value judgment prompts} Experiment 1 used truth value judgment (TVJ) prompts used to probe semantic understanding of aspect in LLMs are in Tables~\ref{tab:tvjsemantic} and~\ref{tab:tvjnarrative}.

\begin{table*}[h]
  \renewcommand{\arraystretch}{1.2} 
  \small
  \begin{tabular}{p{0.9\linewidth}} 
    \texttt{This is a task designed to test your ability to manage two different tasks at once: Task 1: to read and comprehend a sentence, and Task 2: to assess how accurate your evaluation of a phrase is. The phrases are 2-5 words long and are not complete sentences. You should decide *whether the phrase is true* with respect to the original sentence. You should respond with "True" or "False" or "Both" or "Can't Decide" according to the first understanding that comes to mind. Make sure your response contains ONLY your truth value judgment (True/False/Both/Can't Decide). Here's an example:} \vspace{0.5em} \\ 

    \texttt{SENTENCE: Nancy painted her wall} \\
    \texttt{PHRASE: whole wall with fresh paint} \\
    \texttt{Is this phrase true with respect to the sentence?} \\
    \texttt{Option 1 - True} \\
    \texttt{Option 2 - False} \\
    \texttt{Option 3 - Both} \\
    \texttt{Option 4 - Can't Decide} \vspace{0.5em} \\

    \texttt{Response: True} \vspace{0.5em} \\

    \texttt{SENTENCE: \{\{SENTENCE\}\}} \\
    \texttt{PHRASE: \{\{PHRASE\}\}} \\
    \texttt{\{\{OPTIONS\}\}} \vspace{0.5em} \\

    \texttt{Response:} \vspace{0.5em} \\

  \end{tabular}
  \caption{Prompt that was used for TVJ semantic experiment.}\label{tab:tvjsemantic}
\end{table*}


\begin{figure}
    \centering
    \includegraphics[width=0.9\linewidth]{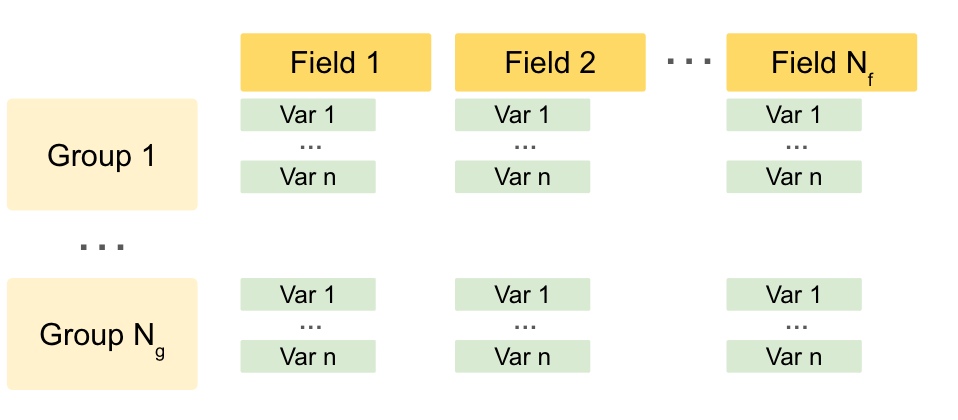}
    \caption{We assume as little structure as possible for experimental datasets to allow for generalizability to other domains. Datasets consist of multiple \textit{groups} of stimuli, which have different independent variable values (users can indicate which \textit{fields} are independent variables). Human studies often compare metrics across different stimuli groups to draw conclusions about the effects of independent variables.}
    \label{fig:datastructure}
\end{figure}

\subsection{Application Usage}
Based on standard experimental procedures in human studies, we structure our expected experiment dataset format such that each row contains an individual stimulus and any number of fields describing that stimulus. Users can then identify which fields correspond to the \textit{independent variables} of their study and define \textit{groups} for their dataset based on those independent variables. (Dataset structure is illustrated in Figure~\ref{fig:datastructure}.) Finally, users can specify \textit{predictions} -- what \textit{dependent variable} should be measured, and how that measure is expected to vary between groups. Users also indicate the dependent variable(s) their study should measure in the LLM response: frequency of responses matching a target value, token probabilities of a target response, or average numerical value of a response (e.g., average rating for experiments that ask for numerical ratings). Instructions are modified for the language model.

\end{document}